\documentclass[11pt, letterpaper]{berkeley}

\usepackage[all]{hypcap}
\usepackage[comma,numbers,sort,compress]{natbib}
\usepackage{hyperref}[citecolor=magenta]

\hypersetup{
    colorlinks = true,
    citecolor = {magenta},
}

\usepackage{microtype}
\usepackage{graphicx}
\usepackage{subfigure}
\usepackage{booktabs}
\usepackage{float}

\usepackage{amsmath}
\usepackage{amssymb}
\usepackage{mathtools}
\usepackage{amsthm}
\usepackage{mathrsfs}
\usepackage{nicefrac}
\usepackage{dsfont}
\usepackage{enumitem}
\usepackage{float}

\usepackage{xspace}
\usepackage[capitalize,noabbrev]{cleveref}
\usepackage{subcaption}
\usepackage{wrapfig}
\usepackage{lipsum}
\usepackage{listings}

\usepackage{amsmath}
\usepackage{amssymb}
\usepackage{mathtools}
\usepackage{amsthm}
\usepackage{bbm}
\usepackage{tabularx}
\usepackage{multirow}
\usepackage[export]{adjustbox}
\usepackage{colortbl}

\usepackage{setspace}

\usepackage{hyperref}
\usepackage{url}
\usepackage{microtype}
\usepackage{amsmath}
\usepackage{amssymb}
\usepackage{bbm}
\usepackage{xcolor}
\usepackage{algorithm}
\usepackage{algpseudocode}
\usepackage{graphicx}
\usepackage{wrapfig}
\usepackage{booktabs}
\usepackage{pifont} 
\usepackage[dvipsnames]{xcolor}

\usepackage{color}
\definecolor{deepblue}{rgb}{0,0,0.5}
\definecolor{deepred}{rgb}{0.6,0,0}
\definecolor{deepgreen}{rgb}{0,0.5,0}
\definecolor{boost_correct_to_correct}{HTML}{66C2A5}
\definecolor{default_correct_to_correct}{HTML}{fc8d62}
\definecolor{dup_correct_to_correct}{HTML}{8da0cb}
\definecolor{new_correct_to_correct}{HTML}{e78ac3}

\newcommand\pythonstyle{\lstset{
basicstyle=\ttfamily\footnotesize,
language=Python,
morekeywords={self, clip, exp, mse_loss, uniform_sample, concatenate, logsumexp},              
keywordstyle=\color{deepblue},
emph={MyClass,__init__},          
emphstyle=\color{deepred},   
stringstyle=\color{deepgreen},
frame=single,                       
showstringspaces=false
}}

\lstnewenvironment{python}[1][]
{
\pythonstyle
\lstset{#1}
}
{}

\newcommand\pythoninline[1]{{\pythonstyle\lstinline!#1!}}

\newcommand{\our}[0]{\texttt{SMART}}
\newcommand{\ourfull}[0]{\texttt{SMART}: Self-learning Meta-strategy Agent for Reasoning Tasks}

\definecolor{blanchedalmond}{rgb}{1.0, 0.92, 0.8}
\definecolor{carmine}{rgb}{0.59, 0.0, 0.09}
\definecolor{lightblue}{rgb}{0.22,0.45,0.70}% light blue

\renewcommand{\mathbf}{\boldsymbol}

\makeatletter
\def\Ddots{\mathinner{\mkern1mu\raise\p@
\vbox{\kern7\p@\hbox{.}}\mkern2mu
\raise4\p@\hbox{.}\mkern2mu\raise7\p@\hbox{.}\mkern1mu}}
\makeatother
\newcommand{\mypar}[1]{\textbf{#1:}}
\numberwithin{equation}{section}

\definecolor{amaranth}{rgb}{0.9, 0.17, 0.31}
\definecolor{antiquebrass}{rgb}{0.8, 0.58, 0.46}
\definecolor{antiquefuchsia}{rgb}{0.57, 0.36, 0.51}
\definecolor{chromeyellow}{rgb}{0.31, 0.47, 0.26}

\usepackage[dvipsnames]{xcolor}
\definecolor{maj5}{HTML}{2b8cbe}
\definecolor{maj5Imp}{HTML}{084081}

\definecolor{seq5wo}{HTML}{d95f0e}
\definecolor{seq5woImp}{HTML}{662506}

\definecolor{seq5w}{HTML}{6a51a3}
\definecolor{seq5wImp}{HTML}{3f007d}

\definecolor{selfwo}{HTML}{d95f0e}
\definecolor{selfwoImp}{HTML}{662506}

\definecolor{selfw}{HTML}{6a51a3}
\definecolor{selfwImp}{HTML}{3f007d}

\definecolor{glorewo}{HTML}{d95f0e}
\definecolor{glorewoImp}{HTML}{662506}

\definecolor{glorew}{HTML}{6a51a3}
\definecolor{glorewImp}{HTML}{3f007d}

\definecolor{vstar}{HTML}{d95f0e}
\definecolor{vstarImp}{HTML}{662506}

\makeatletter
\def\mathcolor#1#{\@mathcolor{#1}}
\def\@mathcolor#1#2#3{%
  \protect\leavevmode
  \begingroup
    \color#1{#2}#3%
  \endgroup
}
\makeatother

\usepackage[textsize=tiny]{todonotes}
\usepackage{amssymb}

\usepackage[skins,theorems]{tcolorbox}
\Crefformat{equation}{#2Eq.\;(#1)#3}

\Crefformat{figure}{#2Figure #1#3}
\Crefformat{assumption}{#2Assumption #1#3}
\Crefname{assumption}{Assumption}{Assumptions}

\usepackage{crossreftools}
\usepackage[suppress]{color-edits}
\usepackage{dsfont}
\usepackage{nicefrac}

\newtcolorbox{analysisbox}[1][]{
    enhanced jigsaw,
    colback=white,
    colframe=blue!75!black,
    fonttitle=\bfseries,
    boxsep=5pt,
    left=5pt,
    right=5pt,
    top=5pt,
    bottom=5pt,
    title=#1,
}

\usepackage{pifont}
\usepackage{soul}
\definecolor{highlightmistake}{RGB}{255, 179, 179}
\definecolor{highlightcorrect}{RGB}{179, 255, 179}

\title{\texttt{SMART}: \textbf{S}elf-learning \textbf{M}eta-strategy \textbf{A}gent for \textbf{R}easoning \textbf{T}asks}
\author[1]{Rongxing Liu*}
\author[1]{Kumar Shridhar*}
\author[1, 2]{Manish Prajapat}
\author[3]{Patrick Xia}
\author[1]{Mrinmaya Sachan}

\affil[1]{ETH Zurich}
\affil[2]{ETH AI Center}
\affil[3]{Microsoft}

\correspondingauthor{\texttt{(*)} equal contribution. Correspondence at: \{\texttt{rongxingtianxia611@163.com, shkumar@ethz.ch}\}  }

\begin{abstract}
Tasks requiring deductive reasoning, especially those involving multiple steps, often demand adaptive strategies such as intermediate generation of rationales or programs, as no single approach is universally optimal. 
While Language Models (LMs) can enhance their outputs through iterative self-refinement and strategy adjustments, they frequently fail to apply the most effective strategy in their first attempt. This inefficiency raises the question: \emph{Can LMs learn to select the optimal strategy in the first attempt, without a need for refinement?}
To address this challenge, we introduce \texttt{SMART} (\textbf{S}elf-learning \textbf{M}eta-strategy \textbf{A}gent for \textbf{R}easoning \textbf{T}asks), a novel framework that enables LMs to autonomously learn and select the most effective strategies for various reasoning tasks. We model the strategy selection process as a \emph{Markov Decision Process} and leverage reinforcement learning-driven continuous self-improvement to allow the model to find the suitable strategy to solve a given task. Unlike traditional self-refinement methods that rely on multiple inference passes or external feedback, \our\ allows an LM to internalize the outcomes of its own reasoning processes and adjust its strategy accordingly, aiming for correct solutions on the first attempt.
Our experiments across various reasoning datasets and with different model architectures demonstrate that \our\ significantly enhances the ability of models to choose optimal strategies without external guidance (+15 points on the GSM8K dataset). By achieving higher accuracy with a single inference pass, \our\ not only improves performance but also reduces computational costs for refinement-based strategies, paving the way for more efficient and intelligent reasoning in LMs.

\vspace{0.25em}
\hspace{0.7em}\includegraphics[width=1.0em,height=1.0em]{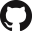}\hspace{.5em}\parbox{\dimexpr\linewidth-2\fboxsep-2\fboxrule}{\url{https://github.com/kumar-shridhar/SMART/}}
\vspace{-.5em}

\end{abstract}

\begin{document}
\maketitle

\section{Introduction}

When people first encounter complex reasoning tasks, such as solving mathematical problems, they often make mistakes or approach them inefficiently \citep{brown2019strategy}. However, with experience, humans tend to improve their performance by replacing ineffective or incorrect strategies with more effective ones, using a mix of strategies tailored to the specific task \citep[][\textit{inter alia}]{adolph1998learning, lemaire2009children, torbeyns2009acquisition, boncoddo2010emergence}.

Language Models (LMs) similarly struggle with reasoning tasks, sometimes producing incoherent results \citep{self-refine, shridhar2023screws, cannotselfrefine}. A common remedy is to resample the output, a process known as \emph{refinement}. This refinement may involve reusing the same reasoning approach \citep{self-refine} or adopting an entirely new one \citep{shridhar2023screws}. In addition, providing feedback on initial results has proven beneficial during resampling \citep[][\textit{inter alia}]{cannotselfrefine, welleck2022generating, shinn2024reflexion, kim2024language}. 
This raises a critical question: \emph{Can LMs be taught to optimize their choice of reasoning strategy for specific tasks overtime on the first trial, much like humans do?}

To address this question, we propose a novel framework called \texttt{SMART} (\textbf{S}elf-learning \textbf{M}eta-strategy \textbf{A}gent for \textbf{R}easoning \textbf{T}asks), which allows LMs to learn optimal strategy selection through a continuous self-learning approach. We model the task of identifying the optimal strategy as a \emph{Markov Decision Process} (MDP) \citep{sutton1999reinforcement, puterman2014markov}, where the agent (LM) starts with its pre-trained knowledge and iteratively improves its performance by learning from its own outputs and strategy choices. By integrating the LM's reasoning abilities with reinforcement learning-driven self-improvement, the agent can simulate different reasoning strategies,  evaluate their effectiveness based on past outcomes, and adjust its strategy choice accordingly.

Our approach differs from traditional methods by focusing on iterative reward-based learning, which encourages the agent to produce the correct inference on the first attempt without resampling. This not only improves cost efficiency - only one sampling step is required during inference - but also results in a more generalizable model capable of adapting its strategy selection based on the specific task. We validate \our\ on a variety of reasoning datasets and LM architectures and show that our method significantly improves the ability of LMs to select optimal strategies on the first try, outperforming baseline models that rely on traditional self-refinement techniques in both accuracy and computational efficiency.
On three mathematical datasets (GSM8K \cite{cobbe2021gsm8k}, SVAMP \cite{svamp}, ASDiv \cite{asdiv}) over three LLM agents (Llama3 8B \cite{llama3}, Gemma 7B \cite{team2024gemma}, and Mistral 7B \cite{jiang2023mistral}), we demonstrate the effectiveness of our approach. On iterative refinement with \our, we achieve gains of up to +15 points (a relative gain of +35\%) on the GSM8K dataset without the need for refinement. In addition, we improve refinement accuracy by +16 points over baselines.

\section{Methodology}

\begin{figure}
    \centering
    \includegraphics[width=1.0\linewidth]{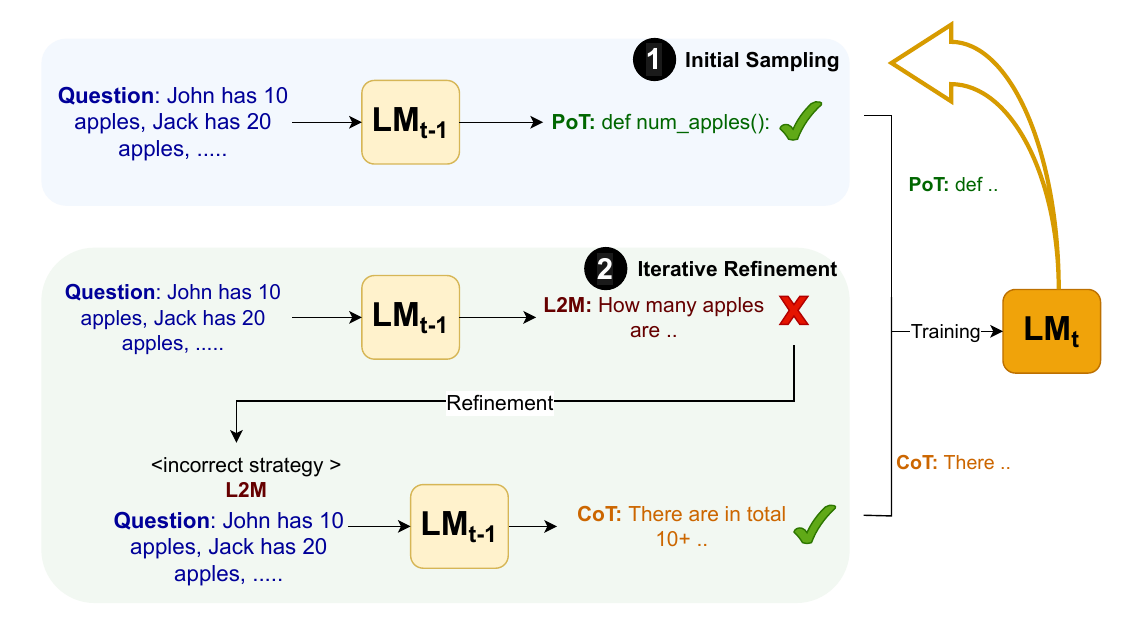}
    \caption{\textbf{Our proposed methodology}: In the first step (initial sampling), an agent (LM) chooses a strategy and solves the given task with it. If it is correct, the process ends successfully. If an incorrect strategy is chosen, the agent iteratively refines its strategy, taking previous strategies into account. The process stops when a correct strategy is chosen to solve a task, or when a stopping criterion such as the number of attempts is reached. All correct strategies are used to further refine the model, and the process is repeated. During testing, we sample once from LM$_t$ without refinement.}
    \label{fig:main-fig}
\end{figure}

Let $q$ be a problem best solved with multi-step reasoning. An example is presented in the form of a mathematical word problem in \autoref{fig:main-fig}. An agent or language model (LM) can approach it using various strategies, such as solving it step by step (Chain of Thought, CoT \cite{cot}), decomposing it into subproblems, and solving each one (Least to Most, L2M \cite{l2m}), or writing a program to solve it programmatically (Program of Thought, PoT \cite{pot}), among others. A common method is to prompt the LM with a specific strategy for solving the task. However, LMs are error-prone but can fix their answers when asked to do so, a process called \emph{refinement}.
In the refinement process, LMs can either stick to the same strategy \cite{self-refine} or switch to a more effective reasoning strategy \cite{shridhar2023screws}. Ideally, LMs could learn to choose the best strategy and reasoning path on the first try, minimizing the need for costly refinement.

\mypar{Our objective}
The primary goal of our work is to enable language models (LMs) to autonomously learn and select the most effective strategies for various reasoning tasks on their first attempt, thereby improving both efficiency and accuracy. Unlike traditional self-refinement methods that require multiple inference passes or external feedback, our approach aims to internalize the learning process within the LM, allowing it to adjust its strategy selection based on past experience. This mirrors how humans learn to choose optimal strategies through experience when faced with complex tasks.

\subsection{\ourfull}

We model the strategy selection process as a \emph{Markov Decision Process} (MDP), where the LM acts as an agent that interacts with the environment (the reasoning tasks) by selecting strategies and observing the outcomes. Using reinforcement learning techniques, the LM can learn a policy that maximizes the expected reward, effectively learning to choose the optimal strategy for each task. This framework allows the LM to simulate different reasoning strategies, evaluate their effectiveness based on past outcomes, and adjust its strategy choice accordingly.

In the following sections, we formalize the problem formulation, define the agent's policy, and describe the learning objective. We then present our two-stage process with iterative refinement, which allows the agent to learn from its own reasoning processes and improve its strategy choice over time.

\mypar{MDP Setup}
We model the strategy selection framework as a Markov Decision Process (MDP) given by the tuple \(\langle \mathcal{S}, \mathcal{A}, \mathcal{P}, \mathcal{R}, \mu \rangle\). Here, $\mathcal{S}$ represents the state space encapsulating all possible states of the environment, where each state \(s \in \mathcal{S}\) is the current problem statement or the subsequent LM response to it. The initial state distribution is given by $\mu$. The action space,
\(\mathcal{A}\), is the set of all strategies available to the agent, where each action \(a_t \in \mathcal{A}\) corresponds to the choice of a particular strategy at time \(t\). The transition function \(\mathcal{P}: \mathcal{S} \times \mathcal{A} \rightarrow \mathcal{S}\) defines the probability of transitioning to a new state \(s_{t+1}\) after applying strategy \(a_t\) in state \(s_t\). 
Particularly for our case, the transition function is non-deterministic as the next state is sampled by the agent (see Algorithm \ref{algo} later).
The reward function \(\mathcal{R}: \mathcal{S} \times \mathcal{A} \rightarrow \mathbb{R}\) assigns a scalar reward based on the correctness of the result after applying the chosen strategy.

We start with the initial sampling step, where the LM chooses a strategy to solve the given task. In other words, given a problem statement $s_1 \sim \mu$, the agent draws a strategy $a_1$ to solve the problem according to its policy (see below). Given the initial problem $s_1$ and strategy $a_1$, the environment transitions to the next state $s_2 \sim \mathcal{P}(\cdot | s_1, a_1)$ with transition probability $\mathcal{P}$ and receives a reward $r_1$ indicating the correctness of the response. If a correct strategy is chosen and the LM solves the task using that strategy, the process terminates. Otherwise, the agent chooses another strategy $a_2$ to solve the problem, and the process repeats until the problem is solved. A realization of this stochastic process is a trajectory $\tau = \left( \left(s_t, a_t, r_t \right)_{t=1}^{T-1}, s_T \right)$ up to a finite number of steps $T$. We denote a partial trajectory (history up to time $t$) as $\tau_{1:t} = \left( \left(s_t, a_t, r_t \right)_{t=1}^{t-1}, s_t \right)$.

\mypar{Agent's policy} We start by defining a history $h_t \coloneqq \tau_{1:t}$ that includes the past actions up to time \mbox{\(t-1\)} and the observed states up to time \(t\). Then, we model the agent using a non-Markovian stochastic policy \(\pi_\theta(a_t | h_t)\) parameterized by \(\theta\), which models the probability of choosing the action \(a_t\) given the history \(h_t\):
\[
\pi_\theta(a_t | h_t) = \Pr(a_t = a \, | \, h_t; \theta)
\]

\mypar{Objective function} We want to optimize the following objective:
\begin{equation}
\centering
\label{objective_fn}
\theta^\star = \arg\max_{\theta} J(\pi_\theta) = \arg\max_{\theta} \mathbb{E}_{s_1, a_1 \sim \mu_{\text{test}}, \pi_{\theta}(a_1|s_1)} [ r_1(s_1,a_1) ]
\end{equation}

where, $\mu_{\text{test}}$ represents the state distribution over the test data.

\subsection*{Two-Stage Process with Refinement}

\subsection*{Stage 1 (Initial Sampling)}

\begin{enumerate}
    \item The process begins with a problem statement $s_1 \sim \mu$ where $\mu$ represents the initial state distribution.
    \item The agent samples an action \(a_1\) from the policy \(\pi_\theta(a_1 | h_1)\), where $h_1 = s_1$ for the step 1.
    \item The agent then generates an output based on strategy \(a_1\)
    \item The agent receives reward \(r_1(h_1,a_1)\) based on it correctness as follows:
    \[
    r_1 = \begin{cases}
    1, & \text{if the output is correct} \\
    0, & \text{otherwise}
    \end{cases}
    \]
    \item \textbf{Termination Check:} If \(r_1 = 1\), the process terminates successfully.
\end{enumerate}

\subsection*{Stage 2 (Iterative Refinement, if \( r_1 = 0 \))}

\textbf{For each time step \(t = 2\) to \(T\):}
    \begin{enumerate}
        \item The agent observes history \(h_t\).
        \item Samples an action \(a_t \sim \pi_\theta(a_t | h_t)\).
        \item Generates an output based on strategy \(a_t\).
        \item Receives reward \(r_t(s_t,a_t)\) :
        \[
        r_t = \begin{cases}
        1, & \text{if the output is correct} \\
        0, & \text{otherwise}
        \end{cases}
        \]
        \item \textbf{Termination Check:} If \(r_t = 1\), terminate the process.
        \item Otherwise, transition to the next state \(s_{t+1}\), which includes the reasoning choice from all the previous steps.
    \end{enumerate}

\mypar{Trajectory and Reward Structure} 
The resulting trajectory \(\tau\) with reward has the following structure:
\[
\tau = \left( (s_1, a_1, r_1), (s_2, a_2, r_2), \ldots, (s_{T'}, a_{T'}, r_{T'}), s_{T'+1} \right)
\]
where \(T' < T\) is the time step at which the process terminates (either by solving the problem or by reaching the maximum steps). We keep the number of trajectories one less than the number of strategies in our work, since each time the model samples with a different strategy than the one present in its history. For simplicity, we omit $r_t=0$ rewards in the trajectory above.

The total reward for the trajectory is:
\[
R(\tau) = \sum_{t=1}^{T'} r_t
\]
Given that \(r_t = 0\) for \(t < T'\) and \(r_{T'} = 1\), if successful, total reward would be \(R(\tau) = 1\). However, in some cases, when a task is never solved, the total reward for that trajectory can be 0. 

As typical in RL, the agent aims to learn a policy that maximizes the expected cumulative reward \citep{sutton1999reinforcement,Prajapat2024RL}:
\begin{equation}
\theta^\star = \arg\max_{\theta} J(\pi_\theta) = \arg\max_{\theta} \mathbb{E}_{\tau \sim f(\tau; \pi_\theta)} [ R(\tau) ]
\end{equation}

\subsection*{Policy Gradient Update}

The gradient of the expected reward with respect to the policy parameters \(\theta\) is:

\begin{equation}
    \nabla_{\theta} J(\pi_\theta) = \mathbb{E}_{\tau} \left[ \sum_{t=1}^{|\tau|-1} \nabla_{\theta} \log \pi_\theta(a_t | h_t) \cdot r_t \right]
\end{equation}

The gradient is computed iteratively, and a step is only taken at time $t$ only if all the previous attempts were incorrect.

\mypar{Implicit bias for Stage 1 to choose the right strategy in the first attempt}  As described in \eqref{objective_fn}, our goal is to maximize the reward received as early as possible. To implicitly bias the model towards selecting the correct action in earlier steps, we adjust the dataset \(\mathcal{D}\) by replacing the sequences of unsuccessful actions with the final successful action taken at the initial state. Specifically, for trajectories where the problem is solved at time \(T'\) (\(T' > 1\)), we replace the samples \((h_1, a_1, r_1), \ldots, (h_{T'}, a_{T'}, r_{T'})\) with \((h_1, a_{T'}, r_{T'})\). This encourages the model to learn to take the correct strategy \(a_{T'}\) for the problem statement \(s_1\) in the first step.

Since we only update the model based on correct outputs, the policy update can be viewed as maximizing the likelihood of the correct actions given the history:
\begin{equation}
\theta^\star = \arg\max_{\theta} \sum_{(h_i, a_i) \in D} \log \pi_\theta(a_i | h_i)
\end{equation}
See Algorithm \ref{algo} for a complete algorithm that illustrates our methodology in detail.

\section*{Algorithm: \ourfull}

\begin{algorithm}[H]
\caption{Training Procedure for \our}
\fontsize{10}{9.5}\selectfont
\label{algo}
\begin{algorithmic}[1]
\Require Initialized policy parameters $\theta$, learning rate $\alpha$, dataset of problems $\mathcal{D} $.
\For{each iteration $e = 1, 2, \hdots$} \Comment{For a fixed number of iterations or until convergence}
\State  \(\mathcal{D}_e \leftarrow \emptyset\)
\Comment{Initialize empty dataset}
\For{each problem \(s_1 \in \mathcal{D}\)}
    \State \textcolor{blue}{Stage 1: Initial Attempt}
    \State Observe initial state $s_1$ 
    \State Sample action $a_1 \sim \pi_\theta(a \mid s_1)$
    \State Generate output using strategy $a_1$
    \State Evaluate output to obtain reward $r_1$
    \If{$r_1 = 1$} \Comment{Correct output}
        \State Collect sample $(s_1, a_1, r_1)$
        \State $\mathcal{D}_e \gets \mathcal{D}_e \cup \{(s_1, a_1)\}$ 
        \Comment{Add sample to dataset}
        \State \textbf{Continue} to next problem
    \Else
        \State \textcolor{red}{Stage 2: Iterative Refinement}
        \For{each refinement iteration $t = 2 \hdots T $} % \Comment{Generally T = number of strategies - 1}
        \State Observe state \(s_t\) and history \(h_t\)
        \State Sample action $a_t \sim \pi_\theta(a | h_t).$
        \State Generate output using strategy $a_t$
        \State Evaluate output to obtain reward $r_t$
        \If{$r_t = 1$} \Comment{Correct output}
            \State Collect sample $(s_1, a_1, \ldots s_t, a_t, r_t)$
            \State $\mathcal{D}_e \gets \mathcal{D}_e \cup \{(s_1, a_1, \ldots s_t, a_t)\}$ 
            \Comment{Add sample to dataset}
            \State \textbf{Break} loop and proceed to next problem
        \EndIf
        \EndFor
    \EndIf
\EndFor
\State \textbf{Policy Update}
\State Update policy parameters $\theta$ using collected data:
\[
    \theta \leftarrow \theta + \alpha \cdot \sum_{(h_i, a_i) \in \mathcal{D}_e} \nabla_{\theta} \log \pi_\theta(a_i | h_i)
    \]
\State \textbf{Implicit Biasing}
\State $\mathcal{D}_{e+1} \leftarrow \mathcal{D}_{e} \backslash \{(s_1, a_1, \ldots s_t, a_t)\} \bigcup \{(s_1,a_t)\}$ \Comment{Update the dataset}
\EndFor
\end{algorithmic}
\end{algorithm}
\section{Experimental Details}

\our{} operates within a \emph{self-learn} framework, where we prompt a pre-trained model with 8-shot examples to collect the initial training data. Prompts used for various strategies are provided in  Subsection \ref{appendix:prompts}. The 8-shot pre-trained model also serves as a baseline comparison for our method. We use the MetaMath dataset \cite{yu2024metamath}, a variant of the GSM8K \cite{cobbe2021gsm8k} training set, which contains 110K reasoning problems. We evaluate our methodology on the GSM8K test set, which contains 1,319 samples. To show the generalization ability of our method, we also test on two out-of-distribution datasets: the SVAMP dataset \cite{svamp} with 1,000 samples where the questions are made more challenging by altering them in a non-trivial way, and the ASDiv dataset \cite{asdiv} with 2,300 samples, which consists of diverse mathematical problems from elementary to middle school.  

\begin{table*}[h!]
\centering %
\small

% \addtolength{\tabcolsep}{-3.5pt}
{ % 
\begin{tabular}{l  l  c c c }
\hline
\multirow{2}{*}{\textbf{Model}} & \multirow{2}{*}{\textbf{Method}} & \multirow{2}{*}{\textbf{Test Accuracy (\%)}} & \multicolumn{2}{c}{\textbf{Refinement}} \\
 &  &  & Strategy & Accuracy (\%) \\
\hline
\hline
& Baseline (\emph{Using 8-shot examples}) & & &   \\
& \quad \multirow{2}{*}{Chain of Thought (CoT)} & \multirow{2}{*}{40.0} & same & 44.6   \\
&  &  & different &  49.4  \\
& \quad \multirow{2}{*}{Least to Most (L2M)} & \multirow{2}{*}{34.9} & same & 43.4 \\
&  &  & different &  48.7  \\
& \quad \multirow{2}{*}{Program of Thought (PoT)} & \multirow{2}{*}{\underline{40.4}} & same & 46.1 \\
&  &  & different & \underline{51.3}  \\
 Gemma 7B & & & \\
& \our\ (\emph{Proposed Approach})  &  & & \\
& \quad \emph{Iteration 1}  & 46.5 & \our & 64.6\\
& \quad \emph{Iteration 2} & 50.6 & \our & 64.0\\
& \quad \emph{Final Iteration - Iteration 5} & \textbf{55.6} \scalebox{0.7}{($\uparrow$ +15.2)} & \our & \textbf{67.5} \scalebox{0.7}{($\uparrow$ +16.2)}\\
\hline
\hline
\\
& Baseline (\emph{Using 8-shot examples}) &  & & \\
& \quad \multirow{2}{*}{Chain of Thought (CoT)} & \multirow{2}{*}{50.6} & same & 59.0\\
&  &  & different &  67.5 \\
& \quad \multirow{2}{*}{Least to Most (L2M)} & \multirow{2}{*}{52.4} & same & 56.6 \\
&  &  & different & 67.2  \\
& \quad \multirow{2}{*}{Program of Thought (PoT)} & \multirow{2}{*}{\underline{56.9}} & same & 61.3\\
&  &  & different &  \underline{70.1}\\
Mistral 7B & & & & \\
& \our\ (\emph{Proposed Approach})  & & &  \\
& \quad \emph{Iteration 1}  & 63.8 & \our & 74.1 \\
& \quad \emph{Iteration 2} & 66.3 & \our & 76.4\\
& \quad \emph{Final Iteration - Iteration 3} & \textbf{67.9} \scalebox{0.7}{($\uparrow$ +11.0)} & \our & \textbf{78.0} \scalebox{0.7}{($\uparrow$ +7.9)} \\
\hline
\hline
\\
& Baseline (\emph{Using 8-shot examples}) &  \\
& \quad \multirow{2}{*}{Chain of Thought (CoT)} & \multirow{2}{*}{\underline{81.9}} & same & \underline{90.4}\\
&  &  & different &  89.3 \\
& \quad \multirow{2}{*}{Least to Most (L2M)} & \multirow{2}{*}{80.0} & same & 84.9\\
&  &  & different &   88.7 \\
& \quad \multirow{2}{*}{Program of Thought (PoT)} & \multirow{2}{*}{76.9} & same & 86.0\\
&  &  & different &  88.6 \\
Qwen2 7B & & \\
& \our\ (\emph{Proposed Approach})  & & &   \\
& \quad \emph{Iteration 1}  & 84.5 &  \our & 91.4\\
& \quad \emph{Iteration 2} & \textbf{85.4} \scalebox{0.7}{($\uparrow$ +3.5)} & \our & \textbf{91.9} \scalebox{0.7}{($\uparrow$ +1.5)}\\
& \quad \emph{Final Iteration - Iteration 3} & 85.2 & \our & 91.2\\
\hline
\hline

\end{tabular}
}
\caption{Test Accuracy (maj@1) comparison between different baselines using three strategies CoT, L2M, PoT with our approach \our\ on the GSM8K dataset. Baselines used 8-shot in-context examples to generate the output. Additionally, we report refinement accuracy obtained through the Oracle verifier for both the baselines and our approach. Refinements for the baselines can follow the \textit{same} strategy as in \cite{self-refine} and a \textit{different} strategy than the initial one as in \cite{shridhar2023screws}. Results are presented for three models: Gemma 7B, Mistral 7B, and Qwen2 7B. The best results among the baseline are \underline{underlined} while the best overall results are in \textbf{bold}.}
\label{tab:main-table}
\end{table*}

We ran all our experiments on models with 7-8 billion parameters, specifically Gemma 7B \cite{team2024gemma}, Mistral 7B \cite{jiang2023mistral}, Qwen2 7B \cite{yang2024qwen2}, and Llama3 8B \cite{llama3}. This is important for our \emph{self-learn} setup, as the model needs a basic understanding of the task to begin the process. Smaller models often have difficulty starting the process, while very large models are too expensive to iterate over multiple times.

We employed three reasoning strategies: Chain of Thought (CoT) \citep{cot}, Least to Most (L2M) \citep{l2m}, and Program of Thought (PoT) \citep{pot} in our work due to their effectiveness on the multi-step reasoning tasks. We take the best out of these strategies as our baseline (\underline{underlined} in \autoref{tab:main-table}). Since we used a different reasoning strategy during refinement following \citet{shridhar2023screws}, the majority of our experiments are limited to two trajectories as the third trajectory would make the remaining strategy the obvious choice. However, we also explore our approach beyond 3 strategies later (in Section \ref{discussion}). 
We report the top-1 accuracy (maj@1) for all experiments, as we want to test the accuracy of the output on the first try. 
We used a temperature of $0.7$ to generate samples at each iteration.
Since the models are already trained on the downstream datasets during pre-training, we chose to train only part of the model and used LoRA \citep{hu2021lora} with rank 16, alpha 32, and a starting learning rate of 2e-4 with decay. We used the Unsloth library \citep{unslothai_unsloth} for training and we did the inference using the VLLM library \citep{vllm}. Finally, we run \our\ for multiple iterations until our accuracy gains no longer justify the cost of training and sampling. In our experiments, 3 iterations worked well for most models, although we trained Gemma 7B for 5 iterations before performance plateaued. 

\begin{table*}[h!]
\centering %
\small
\caption{Test accuracy (maj@1) comparison between different baselines using three strategies CoT, L2M, PoT with our proposed approach \our\ on the out-of-domain datasets of SVAMP and ASDiv. Baselines used 8-shot in-context examples to generate the output. Results are presented for three models: Gemma 7B, Mistral 7B, and Qwen2 7B. The best baseline results are \underline{underlined}, while the best overall results are in \textbf{bold}. \\} 
% \addtolength{\tabcolsep}{-3.5pt}
{ % 
\begin{tabular}{l  l  c c }
\hline
\textbf{Model} &\textbf{Method} & \textbf{SVAMP Acc (in \%)} & \textbf{ASDiv Acc (in \%)} \\
\hline
\hline
& Baseline (\emph{Using 8-shot examples}) & &    \\
& \quad Chain of Thought (CoT) &  \underline{65.6} & \underline{68.0}    \\
& \quad Least to Most (L2M) &  63.4  & 62.7  \\
Gemma 7B & \quad Program of Thought (PoT)  & 51.6 &  54.2\\
& & & \\
& \our\ (\emph{Proposed Approach})  &  &  \\
& \quad \emph{Final Iteration} & \textbf{68.2} \scalebox{0.7}{($\uparrow$ +2.6)}  & \textbf{70.9} \scalebox{0.7}{($\uparrow$ +2.9)}\\
\hline
\hline
& Baseline (\emph{Using 8-shot examples}) & &    \\
& \quad Chain of Thought (CoT) &  65.8  & \underline{71.3}    \\
& \quad Least to Most (L2M) & 67.9   & 67.3 \\
Mistral 7B & \quad Program of Thought (PoT)  &  \textbf{\underline{71.7}} & 70.1 \\
& & & \\
& \our\ (\emph{Proposed Approach})  &  &  \\
& \quad \emph{Final Iteration} & 70.9 \scalebox{0.7}{($\downarrow$ -0.8)}  & \textbf{72.3} \scalebox{0.7}{($\uparrow$ +1.0)}\\
\hline
\hline
& Baseline (\emph{Using 8-shot examples}) & &    \\
& \quad Chain of Thought (CoT) & \underline{89.3}  & \underline{89.5}    \\
& \quad Least to Most (L2M) &  87.5  & 84.2  \\
Qwen2 7B & \quad Program of Thought (PoT)  & 82.8 & 80.8 \\
& & & \\
& \our\ (\emph{Proposed Approach})  &  &  \\
& \quad \emph{Final Iteration} & \textbf{90.3} \scalebox{0.7}{($\uparrow$ +1.0)}  & \textbf{90.5} \scalebox{0.7}{($\uparrow$ +1.0)}\\
\hline
\hline
\end{tabular}
}
\label{tab:out-of-domain}
\end{table*}

\section{Results}

\mypar{\our\ significantly improves results on in-distribution dataset}
We compared \our\ with baselines on the GSM8K dataset, which we also consider to be an in-distribution dataset since the train and test sets have the same distribution. \autoref{tab:main-table} shows that \our\ outperformed the baseline in its first iteration on the GSM8K dataset, achieving a gain of +6 points for both the Gemma 7B and Mistral 7B models (40.4 $\rightarrow$ 46.5 and 56.9 $\rightarrow$ 63.8, respectively). Although Qwen2's performance is already very strong on the GSM8K dataset, we still observed a gain of +2.6 points (81.9 $\rightarrow$ 84.5) in the first iteration. After a few more iterations, we saw a total gain of +15 points for Gemma 7B (40.4 $\rightarrow$ 55.4), +11 points for Mistral 7B (56.9 $\rightarrow$ 67.9), and +4 points for Qwen2 7B (81.9 $\rightarrow$ 85.4).

\mypar{\our{} serves as a great refinement strategy}
Since \our\ involves iterative refinement to find the optimal action given the trajectory, we also compare against two refinement baselines: refinement with the same strategy \citep{self-refine} and refinement with a strategy change \citep{shridhar2023screws}. We used the Oracle Verifier, which identifies incorrect samples and refines them either with the same strategy or by choosing a different one. \footnote{Note that the Oracle Verifier is used to set the upper bound for all the methods compared and cannot be used during inference. It is only to compare the capabilities of different approaches.} \autoref{tab:main-table} compares the refinement accuracy with \our\, and our proposed methodology shows significant improvements over the baselines. Gemma 7B gains over +16 points (48.9 $\rightarrow$ 67.5) compared to the best refinement baseline, Mistral 7B gains +8 points (66.5 $\rightarrow$ 78.0), and Qwen2 7B gains +1.5 points (86.9 $\rightarrow$ 91.9).

\mypar{\our{} generalizes well to out-of-distribution dataset} 
We test our trained checkpoints using our proposed approach \our\ on two out-of-distribution datasets: ASDiv and SVAMP. These are referred to as out-of-distribution because the model was not trained on these datasets but only evaluated on them. \autoref{tab:out-of-domain} compares the results of \our\ with the baselines (created using 8-shot in context examples) across the three reasoning strategies: CoT, L2M, and PoT. Among the three models, Gemma 7B showed the most improvement, with a gain of +2.6 points on the SVAMP dataset (65.6 $\rightarrow$ 68.2) and +2.9 points on the ASDiv dataset (68.0 $\rightarrow$ 70.9). Mistral 7B, on the other hand, gained +1 point on the ASDiv dataset (71.3 $\rightarrow$ 72.3) but performed worse on the SVAMP dataset. We suspect that the test dataset for SVAMP is different from GSM8K, and since the model was not optimized for SVAMP-style questions, this could explain the drop in performance. Finally, for the Qwen2 7B model, we observed a modest gain of +1 point across both datasets (89.3 $\rightarrow$ 90.3 on SVAMP and 89.5 $\rightarrow$ 90.5 on ASDiv).
\section{Discussion}
\label{discussion}
\begin{figure}
    \centering
    \includegraphics[width=1.0\linewidth]{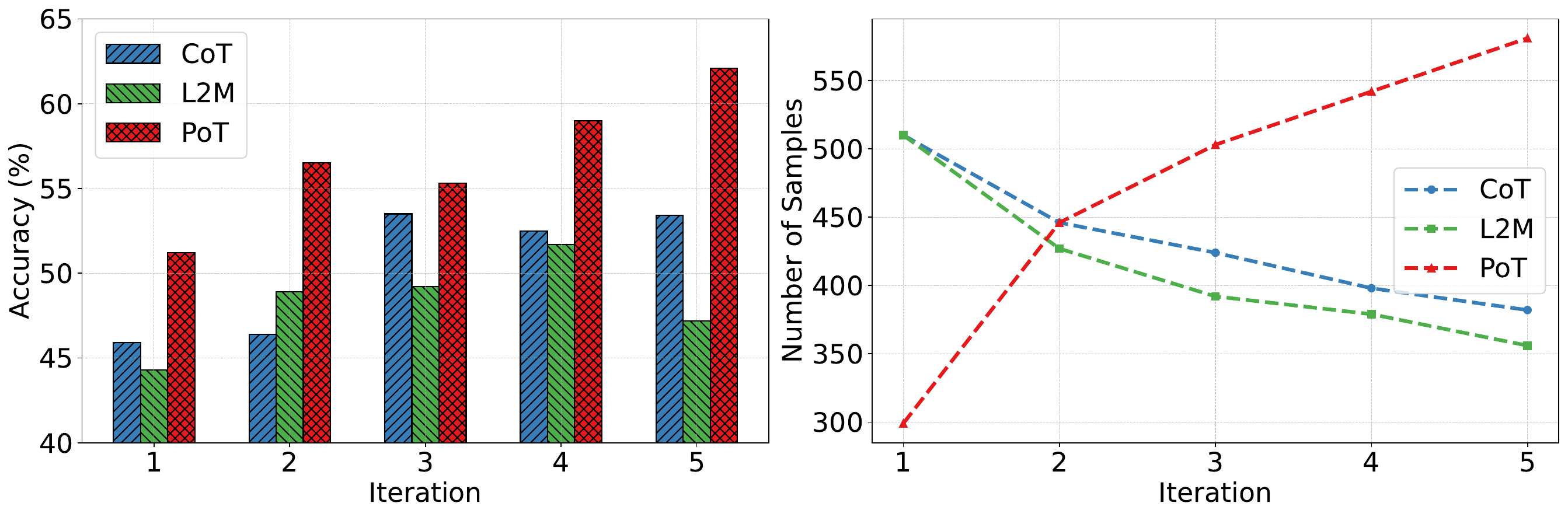}
    \caption{Strategy distribution change over iterations for Gemma 7B model on GSM8K dataset.}
    \label{fig:gemma-str-dist}
\end{figure}

\mypar{How the strategy distribution changes over iterations} With \our, the goal is to select the desired strategy at the first attempt, i.e., over iterations the LM should learn to select the appropriate strategy for a given task. 
\autoref{fig:gemma-str-dist} shows the changes in strategy distribution over iterations for the Gemma 7B model on the GSM8K dataset. As indicated by the baseline in \autoref{tab:main-table}, PoT emerges as the best strategy for the Gemma 7B model, followed by CoT, with L2M being the least effective. A similar pattern is observed in \autoref{fig:gemma-str-dist}, where the model increasingly favors PoT (indicated by the upward trend in the red line) while decreasing its preference for the other two strategies. Correspondingly, the accuracy for PoT improves the most, followed by CoT and L2M, demonstrating that over iterations the model learns to select the optimal strategy for a given task. A qualitative example where over iterations the model learned to choose the right strategy in its first attempt is provided in \autoref{fig:qual-results}. Initially, the model chose the wrong strategy but was fixed during refinement, and over iterations, the model picked the correct strategy in its initial sampling, without the need for refinement. 

\mypar{Extending the strategies beyond three strategies in \our}
\label{extend-strategies}
We want to test if it is possible to go beyond the three strategies explored in this paper by extending our approach to other strategies. Although we could not find a strategy with performance comparable to those explored in this paper, we introduced a \texttt{<Unsolvable>} tag for questions that the model could not answer during either sampling or refinement attempts and tried to use it as a fourth strategy. However, extending this strategy did not yield promising results due to the limited number of samples in the \texttt{<Unsolvable>} category, as the model was able to solve the task using one of the strategies during refinement. Similarly, we experimented with a \texttt{<Answer Only>} strategy, where the model directly predicts the answer to very simple questions without any intermediate reasoning. As with the \texttt{<Unsolvable>} strategy, the skewed data distribution led to poorer performance.

\mypar{Is the strategy selection effective?}
We compare the strategy selection made by \our\ during inference with a fixed strategy (CoT) on the GSM8K dataset using the Gemma 7B model, as shown in \autoref{tab:fixed-str-vs-our}. Across all five iterations, the results highlight the importance of selecting the appropriate strategy, with the \our\ approach outperforming the fixed strategy (CoT) by up to 6 points when the optimal strategy is selected. While the fixed strategy shows improvements over iterations, illustrating the \emph{self-learning} effect over multiple generations, our work demonstrates that performance can be further improved by selecting the most appropriate strategy for each task.

\begin{table}[t]
    \centering
    \caption{Comparison of fixed strategy (CoT) vs \our\ based strategy during inference for GSM8K dataset using Gemma 7B model. \\}
    \label{tab:fixed-str-vs-our}
    \begin{tabular}{l c c c c c}
    \toprule
         \textbf{Strategy} & \textbf{Iteration 1} & \textbf{Iteration 2} & \textbf{Iteration 3} & \textbf{Iteration 4} & \textbf{Iteration 5} \\
         \midrule
         Fixed (CoT) & 45.5 & 47.5 & 48.5 & 50.6 & 49.7\\
         \our\ & \textbf{46.5} & \textbf{50.6} & \textbf{52.9} & \textbf{55.0} & \textbf{55.6} \\ 
        \bottomrule
    \end{tabular}
\end{table}

\mypar{\our\ with better starting samples}

\begin{wrapfigure}{r}{0.55\textwidth}  % 'r' for right, 'l' for left
    \centering
    \vspace{-0.2cm}
    \includegraphics[width=0.55\textwidth]{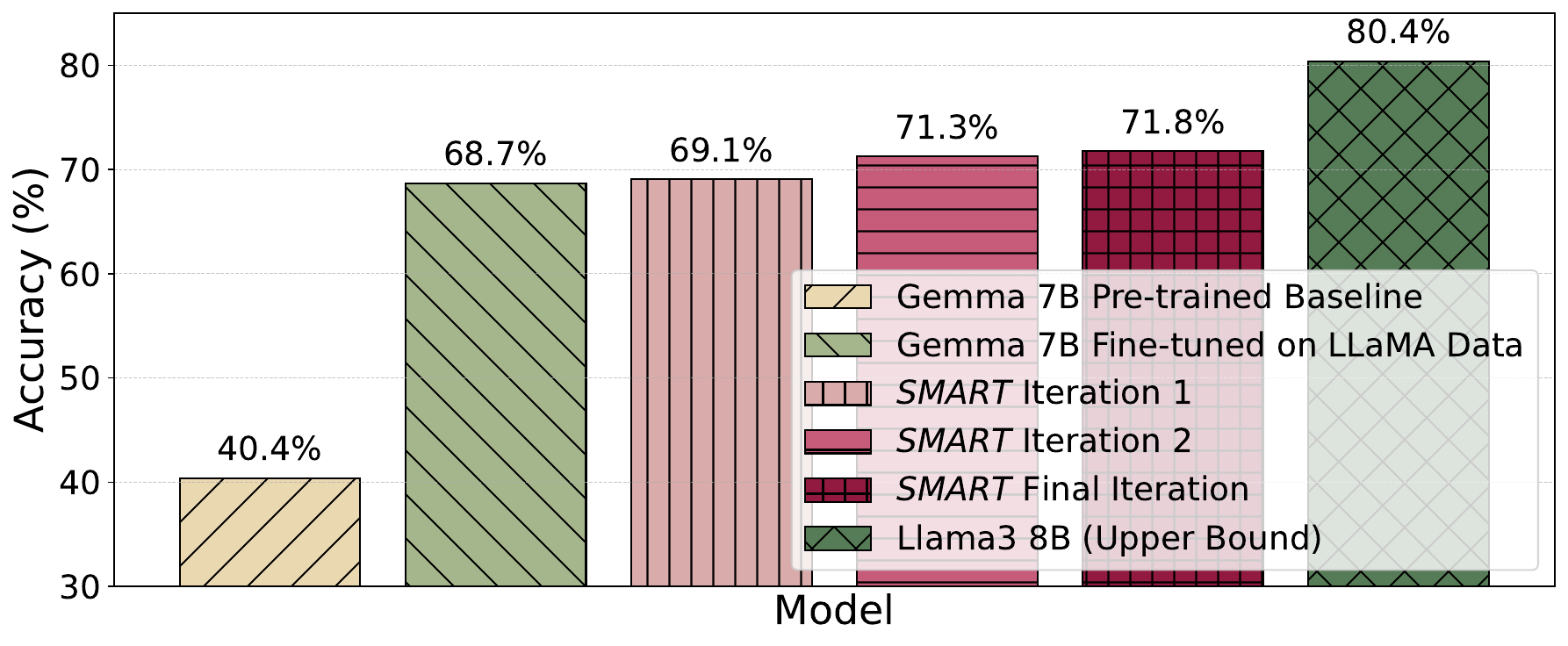}
    \caption{Figure showing a comparison of the effects of different starting data points for the Gemma 7B model. \our\ is compared against two baselines: the pre-trained Gemma 7B model and the fine-tuned Gemma 7B model on Llama3 8B data.}
    \label{fig:smart-gemma-llama}
\end{wrapfigure}

Since we start with the model-generated samples, for a weaker model the starting samples can be improved if those samples come from the stronger model. 
We investigated whether the \our\ based \emph{self-learning} approach could be extended to a setup where the initial data points are collected from a stronger model to initiate training for a weaker model. This is particularly beneficial for weaker models that cannot independently initiate the self-learning process using \our\ due to their limited capabilities on a given task. Initially, we collected data using 8-shot in-context examples generated by the Llama3 8B model \citep{llama3}. This approach yielded an average accuracy of about 80\% across all three strategies, significantly higher than Gemma 7B's initial accuracy of about 40\%. \autoref{fig:smart-gemma-llama} illustrates the comparison between different iterations of \our\ and the pre-trained Gemma 7B baseline at 40.4\%, as well as the fine-tuned Gemma 7B baseline at 68.7\%. Using \our\ resulted in an additional improvement of +3.1 points (from 68.7\% $\rightarrow$ 71.8\%).
\section{Related Work}

\mypar{Refinement in LLMs}
Refinement refers to the process of improving the initial output of large language models (LLMs) through iterative adjustments. This refinement can be achieved by following the same method used initially \citep{self-refine}, by incorporating feedback while using the same approach \citep{welleck2022generating, shinn2024reflexion, kim2024language, cannotselfrefine}, or by using an alternative method \citep{shridhar2023screws, shridhar-etal-2024-art}. However, recent studies have shown that naively applying self-correction can sometimes degrade performance \citep{cannotselfrefine, qu2024recursive, tyen-etal-2024-llms}, highlighting the need for more effective strategies. Supervised fine-tuning with feedback from larger models \citep{ye2023selfee, qu2024recursive}, or using an ensemble of models \cite{havrilla2024glore}, has produced notable results. Nevertheless, relying on larger or multiple models for feedback presents challenges. In contrast, our method learns the optimal strategy during training and applies it correctly on the first attempt, eliminating the need for refinement.

\mypar{Self-Training in LLMs} 
Self-training is a semi-supervised learning method in which the model's own predictions are used as additional data to improve its performance \citep{scudder1965probability, yarowsky-1995-unsupervised}. This technique has been applied to NLP tasks, such as machine translation \citep{He2020Revisiting, sun-etal-2021-self, gulcehre2023reinforced}. We use self-learning principles to generate new data and continually update the policy in an on-policy fashion. This approach can be viewed as an on-policy counterpart to self-imitation learning \citep{sil}, where the policy learns from prospective successful trajectories in its initial stages, rather than imitating past successful behavior.
\section{Conclusion}

\begin{figure*}[t!]
\small
% \sffamily
\centering
\begin{tabular}{l}
\toprule
\addlinespace
\addlinespace
\parbox{\textwidth}{\textbf{Question} 
Brian's friend Bobby has 5 fewer than 3 times as many video games as Brian does.  If Brian has 20 video games but lost 5 right before the comparison was made, how many does Bobby have?} \\
\addlinespace
\parbox{\textwidth}{\textbf{Iteration 1 Stage 1 (Sampling)} \\ 
Program of Thought: brian\_games = 20 \\
games\_lost = 5 \\
bobby\_games = 3 * brian\_games - games\_lost - 5 \\
answer = bobby\_games  
\textcolor{red}{\ding{55}}} \\
\addlinespace
\parbox{\textwidth}{\textbf{Iteration 1 Stage 2 (Refinement)} \\ 
Chain of Thought: Brian has 20 video games and lost 5 games, so he has 20 - 5 = 15 games left. \\
Bobby has 5 fewer than 3 times as many video games as Brian does, so he has 3 * 15 - 5 = 40 video games. \\
Final Answer: 40
\textcolor{green}{\ding{51}}} \\
\addlinespace
\parbox{\textwidth}{\textbf{Iteration 5 Stage 1 (Sampling)} \\ 
Chain of Thought: Brian has 20 - 5 = 15 video games. \\
Bobby has 5 fewer than 3 times as many video games as Brian does, so Bobby has 3 * 15 - 5 = 40 video games. \\
Final Answer: 40
\textcolor{green}{\ding{51}}} \\
\addlinespace
\bottomrule
\end{tabular}
\caption{Qualitative example demonstrating that Gemma 7B model learnt to refinement strategy in its initial sampling stage, removing the need for refinement.}
\label{fig:qual-results}
\end{figure*}

We present \ourfull, a solution to the challenges LMs face in selecting strategies for complex reasoning tasks. By modeling the strategy selection process as a Markov decision process and leveraging reinforcement learning, \texttt{SMART} enables LMs to autonomously learn and apply the most effective reasoning strategies on the first trial, thereby reducing the reliance on iterative self-refinement. Our proposed approach not only improves the accuracy of LMs, as demonstrated by significant performance gains across multiple datasets but also enhances computational efficiency by minimizing the need for multiple inference passes.

\bibliography{main}

\begin{thebibliography}{39}
\providecommand{\natexlab}[1]{#1}
\providecommand{\url}[1]{\texttt{#1}}
\expandafter\ifx\csname urlstyle\endcsname\relax
  \providecommand{\doi}[1]{doi: #1}\else
  \providecommand{\doi}{doi: \begingroup \urlstyle{rm}\Url}\fi

\bibitem[Adolph et~al.(1998)Adolph, Vereijken, and Denny]{adolph1998learning}
Karen~E Adolph, Beatrix Vereijken, and Mark~A Denny.
\newblock Learning to crawl.
\newblock \emph{Child development}, 69\penalty0 (5):\penalty0 1299--1312, 1998.
\newblock URL \url{https://www.jstor.org/stable/1166199}.

\bibitem[Boncoddo et~al.(2010)Boncoddo, Dixon, and Kelley]{boncoddo2010emergence}
Rebecca Boncoddo, James~A Dixon, and Elizabeth Kelley.
\newblock The emergence of a novel representation from action: Evidence from preschoolers.
\newblock \emph{Developmental science}, 13\penalty0 (2):\penalty0 370--377, 2010.
\newblock URL \url{https://pubmed.ncbi.nlm.nih.gov/20136934/}.

\bibitem[Brown et~al.(2019)Brown, Menendez, and Alibali]{brown2019strategy}
Sarah~A Brown, David Menendez, and Martha~W Alibali.
\newblock Strategy adoption depends on characteristics of the instruction, learner, and strategy.
\newblock \emph{Cognitive Research: Principles and Implications}, 4\penalty0 (1):\penalty0 9, 2019.
\newblock URL \url{https://cognitiveresearchjournal.springeropen.com/articles/10.1186/s41235-019-0158-3}.

\bibitem[Chen et~al.(2023)Chen, Ma, Wang, and Cohen]{pot}
Wenhu Chen, Xueguang Ma, Xinyi Wang, and William~W. Cohen.
\newblock Program of thoughts prompting: Disentangling computation from reasoning for numerical reasoning tasks.
\newblock \emph{Transactions on Machine Learning Research}, 2023.
\newblock ISSN 2835-8856.
\newblock URL \url{https://openreview.net/forum?id=YfZ4ZPt8zd}.

\bibitem[Cobbe et~al.(2021)Cobbe, Kosaraju, Bavarian, Chen, Jun, Kaiser, Plappert, Tworek, Hilton, Nakano, et~al.]{cobbe2021gsm8k}
Karl Cobbe, Vineet Kosaraju, Mohammad Bavarian, Mark Chen, Heewoo Jun, Lukasz Kaiser, Matthias Plappert, Jerry Tworek, Jacob Hilton, Reiichiro Nakano, et~al.
\newblock Training verifiers to solve math word problems.
\newblock \emph{arXiv preprint}, 2021.
\newblock URL \url{https://arxiv.org/abs/2110.14168}.

\bibitem[Dubey et~al.(2024)Dubey, Jauhri, Pandey, Kadian, Al-Dahle, Letman, Mathur, Schelten, Yang, Fan, et~al.]{llama3}
Abhimanyu Dubey, Abhinav Jauhri, Abhinav Pandey, Abhishek Kadian, Ahmad Al-Dahle, Aiesha Letman, Akhil Mathur, Alan Schelten, Amy Yang, Angela Fan, et~al.
\newblock The llama 3 herd of models.
\newblock \emph{arXiv preprint}, 2024.
\newblock URL \url{https://arxiv.org/abs/2407.21783}.

\bibitem[Gulcehre et~al.(2023)Gulcehre, Paine, Srinivasan, Konyushkova, Weerts, Sharma, Siddhant, Ahern, Wang, Gu, et~al.]{gulcehre2023reinforced}
Caglar Gulcehre, Tom~Le Paine, Srivatsan Srinivasan, Ksenia Konyushkova, Lotte Weerts, Abhishek Sharma, Aditya Siddhant, Alex Ahern, Miaosen Wang, Chenjie Gu, et~al.
\newblock Reinforced self-training (rest) for language modeling.
\newblock \emph{arXiv preprint}, 2023.
\newblock URL \url{https://arxiv.org/abs/2308.08998}.

\bibitem[Havrilla et~al.(2024)Havrilla, Raparthy, Nalmpantis, Dwivedi-Yu, Zhuravinskyi, Hambro, and Raileanu]{havrilla2024glore}
Alexander Havrilla, Sharath~Chandra Raparthy, Christoforos Nalmpantis, Jane Dwivedi-Yu, Maksym Zhuravinskyi, Eric Hambro, and Roberta Raileanu.
\newblock {GL}o{R}e: When, where, and how to improve {LLM} reasoning via global and local refinements.
\newblock In \emph{Proceedings of the 41st International Conference on Machine Learning}, Proceedings of Machine Learning Research, pages 17719--17733. PMLR, 21--27 Jul 2024.
\newblock URL \url{https://proceedings.mlr.press/v235/havrilla24a.html}.

\bibitem[He et~al.(2020)He, Gu, Shen, and Ranzato]{He2020Revisiting}
Junxian He, Jiatao Gu, Jiajun Shen, and Marc'Aurelio Ranzato.
\newblock Revisiting self-training for neural sequence generation.
\newblock In \emph{International Conference on Learning Representations}, 2020.
\newblock URL \url{https://openreview.net/forum?id=SJgdnAVKDH}.

\bibitem[Hu et~al.(2022)Hu, yelong shen, Wallis, Allen-Zhu, Li, Wang, Wang, and Chen]{hu2021lora}
Edward~J Hu, yelong shen, Phillip Wallis, Zeyuan Allen-Zhu, Yuanzhi Li, Shean Wang, Lu~Wang, and Weizhu Chen.
\newblock Lo{RA}: Low-rank adaptation of large language models.
\newblock In \emph{International Conference on Learning Representations}, 2022.
\newblock URL \url{https://openreview.net/forum?id=nZeVKeeFYf9}.

\bibitem[Huang et~al.(2024)Huang, Chen, Mishra, Zheng, Yu, Song, and Zhou]{cannotselfrefine}
Jie Huang, Xinyun Chen, Swaroop Mishra, Huaixiu~Steven Zheng, Adams~Wei Yu, Xinying Song, and Denny Zhou.
\newblock Large language models cannot self-correct reasoning yet.
\newblock In \emph{The Twelfth International Conference on Learning Representations}, 2024.
\newblock URL \url{https://openreview.net/forum?id=IkmD3fKBPQ}.

\bibitem[Jiang et~al.(2023)Jiang, Sablayrolles, Mensch, Bamford, Chaplot, de~las Casas, Bressand, Lengyel, Lample, Saulnier, et~al.]{jiang2023mistral}
AQ~Jiang, A~Sablayrolles, A~Mensch, C~Bamford, DS~Chaplot, D~de~las Casas, F~Bressand, G~Lengyel, G~Lample, L~Saulnier, et~al.
\newblock Mistral 7b (2023).
\newblock \emph{arXiv preprint}, 2023.
\newblock URL \url{https://arxiv.org/abs/2310.06825}.

\bibitem[Kim et~al.(2023)Kim, Baldi, and McAleer]{kim2024language}
Geunwoo Kim, Pierre Baldi, and Stephen~Marcus McAleer.
\newblock Language models can solve computer tasks.
\newblock In \emph{Thirty-seventh Conference on Neural Information Processing Systems}, 2023.
\newblock URL \url{https://openreview.net/forum?id=M6OmjAZ4CX}.

\bibitem[Kwon et~al.(2023)Kwon, Li, Zhuang, Sheng, Zheng, Yu, Gonzalez, Zhang, and Stoica]{vllm}
Woosuk Kwon, Zhuohan Li, Siyuan Zhuang, Ying Sheng, Lianmin Zheng, Cody~Hao Yu, Joseph Gonzalez, Hao Zhang, and Ion Stoica.
\newblock Efficient memory management for large language model serving with pagedattention.
\newblock In \emph{Proceedings of the 29th Symposium on Operating Systems Principles}. Association for Computing Machinery, 2023.
\newblock URL \url{https://doi.org/10.1145/3600006.3613165}.

\bibitem[Lemaire and Callies(2009)]{lemaire2009children}
Patrick Lemaire and Sophie Callies.
\newblock Children’s strategies in complex arithmetic.
\newblock \emph{Journal of Experimental Child Psychology}, 2009.
\newblock URL \url{https://www.sciencedirect.com/science/article/pii/S0022096508001458}.

\bibitem[Madaan et~al.(2023)Madaan, Tandon, Gupta, Hallinan, Gao, Wiegreffe, Alon, Dziri, Prabhumoye, Yang, Gupta, Majumder, Hermann, Welleck, Yazdanbakhsh, and Clark]{self-refine}
Aman Madaan, Niket Tandon, Prakhar Gupta, Skyler Hallinan, Luyu Gao, Sarah Wiegreffe, Uri Alon, Nouha Dziri, Shrimai Prabhumoye, Yiming Yang, Shashank Gupta, Bodhisattwa~Prasad Majumder, Katherine Hermann, Sean Welleck, Amir Yazdanbakhsh, and Peter Clark.
\newblock Self-refine: Iterative refinement with self-feedback.
\newblock In \emph{Thirty-seventh Conference on Neural Information Processing Systems}, 2023.
\newblock URL \url{https://openreview.net/forum?id=S37hOerQLB}.

\bibitem[Miao et~al.(2020)Miao, Liang, and Su]{asdiv}
Shen-yun Miao, Chao-Chun Liang, and Keh-Yih Su.
\newblock A diverse corpus for evaluating and developing {E}nglish math word problem solvers.
\newblock In \emph{Proceedings of the 58th Annual Meeting of the Association for Computational Linguistics}. Association for Computational Linguistics, July 2020.
\newblock URL \url{https://aclanthology.org/2020.acl-main.92}.

\bibitem[Oh et~al.(2018)Oh, Guo, Singh, and Lee]{sil}
Junhyuk Oh, Yijie Guo, Satinder Singh, and Honglak Lee.
\newblock Self-imitation learning.
\newblock In Jennifer Dy and Andreas Krause, editors, \emph{Proceedings of the 35th International Conference on Machine Learning}, volume~80 of \emph{Proceedings of Machine Learning Research}, pages 3878--3887. PMLR, 10--15 Jul 2018.
\newblock URL \url{https://proceedings.mlr.press/v80/oh18b.html}.

\bibitem[Patel et~al.(2021)Patel, Bhattamishra, and Goyal]{svamp}
Arkil Patel, Satwik Bhattamishra, and Navin Goyal.
\newblock Are {NLP} models really able to solve simple math word problems?
\newblock In \emph{Proceedings of the 2021 Conference of the North American Chapter of the Association for Computational Linguistics: Human Language Technologies}. Association for Computational Linguistics, June 2021.
\newblock URL \url{https://aclanthology.org/2021.naacl-main.168}.

\bibitem[Prajapat et~al.(2024)Prajapat, Mutny, Zeilinger, and Krause]{Prajapat2024RL}
Manish Prajapat, Mojmir Mutny, Melanie~N. Zeilinger, and Andreas Krause.
\newblock Submodular reinforcement learning.
\newblock In \emph{The Twelfth International Conference on Learning Representations, {ICLR} 2024, Vienna, Austria, May 7-11, 2024}. OpenReview.net, 2024.
\newblock URL \url{https://openreview.net/forum?id=loYSzjSaAK}.

\bibitem[Puterman(2014)]{puterman2014markov}
Martin~L Puterman.
\newblock \emph{Markov decision processes: discrete stochastic dynamic programming}.
\newblock John Wiley \& Sons, 2014.

\bibitem[Qu et~al.(2024)Qu, Zhang, Garg, and Kumar]{qu2024recursive}
Yuxiao Qu, Tianjun Zhang, Naman Garg, and Aviral Kumar.
\newblock Recursive introspection: Teaching language model agents how to self-improve.
\newblock \emph{arXiv preprint}, 2024.
\newblock URL \url{https://arxiv.org/abs/2407.18219}.

\bibitem[Scudder(1965)]{scudder1965probability}
H.~Scudder.
\newblock Probability of error of some adaptive pattern-recognition machines.
\newblock \emph{IEEE Transactions on Information Theory}, 11\penalty0 (3):\penalty0 363--371, 1965.
\newblock \doi{10.1109/TIT.1965.1053799}.

\bibitem[Shinn et~al.(2023)Shinn, Cassano, Gopinath, Narasimhan, and Yao]{shinn2024reflexion}
Noah Shinn, Federico Cassano, Ashwin Gopinath, Karthik~R Narasimhan, and Shunyu Yao.
\newblock Reflexion: language agents with verbal reinforcement learning.
\newblock In \emph{Thirty-seventh Conference on Neural Information Processing Systems}, 2023.
\newblock URL \url{https://openreview.net/forum?id=vAElhFcKW6}.

\bibitem[Shridhar et~al.(2023)Shridhar, Jhamtani, Fang, Van~Durme, Eisner, and Xia]{shridhar2023screws}
Kumar Shridhar, Harsh Jhamtani, Hao Fang, Benjamin Van~Durme, Jason Eisner, and Patrick Xia.
\newblock Screws: A modular framework for reasoning with revisions.
\newblock \emph{arXiv preprint}, 2023.
\newblock URL \url{https://arxiv.org/abs/2309.13075}.

\bibitem[Shridhar et~al.(2024)Shridhar, Sinha, Cohen, Wang, Yu, Pasunuru, Sachan, Weston, and Celikyilmaz]{shridhar-etal-2024-art}
Kumar Shridhar, Koustuv Sinha, Andrew Cohen, Tianlu Wang, Ping Yu, Ramakanth Pasunuru, Mrinmaya Sachan, Jason Weston, and Asli Celikyilmaz.
\newblock The {ART} of {LLM} refinement: Ask, refine, and trust.
\newblock In \emph{Proceedings of the 2024 Conference of the North American Chapter of the Association for Computational Linguistics: Human Language Technologies (Volume 1: Long Papers)}. Association for Computational Linguistics, June 2024.
\newblock URL \url{https://aclanthology.org/2024.naacl-long.327}.

\bibitem[Sun et~al.(2021)Sun, Wang, Chen, Utiyama, Sumita, and Zhao]{sun-etal-2021-self}
Haipeng Sun, Rui Wang, Kehai Chen, Masao Utiyama, Eiichiro Sumita, and Tiejun Zhao.
\newblock Self-training for unsupervised neural machine translation in unbalanced training data scenarios.
\newblock In \emph{Proceedings of the 2021 Conference of the North American Chapter of the Association for Computational Linguistics: Human Language Technologies}. Association for Computational Linguistics, June 2021.
\newblock URL \url{https://aclanthology.org/2021.naacl-main.311}.

\bibitem[Sutton et~al.(1999)Sutton, Barto, et~al.]{sutton1999reinforcement}
Richard~S Sutton, Andrew~G Barto, et~al.
\newblock Reinforcement learning.
\newblock \emph{Journal of Cognitive Neuroscience}, 11\penalty0 (1):\penalty0 126--134, 1999.

\bibitem[Team et~al.(2024)Team, Mesnard, Hardin, Dadashi, Bhupatiraju, Pathak, Sifre, Rivi{\`e}re, Kale, Love, et~al.]{team2024gemma}
Gemma Team, Thomas Mesnard, Cassidy Hardin, Robert Dadashi, Surya Bhupatiraju, Shreya Pathak, Laurent Sifre, Morgane Rivi{\`e}re, Mihir~Sanjay Kale, Juliette Love, et~al.
\newblock Gemma: Open models based on gemini research and technology.
\newblock \emph{arXiv preprint}, 2024.
\newblock URL \url{https://arxiv.org/abs/2403.08295}.

\bibitem[Torbeyns et~al.(2009)Torbeyns, De~Smedt, Ghesqui{\`e}re, and Verschaffel]{torbeyns2009acquisition}
Joke Torbeyns, Bert De~Smedt, Pol Ghesqui{\`e}re, and Lieven Verschaffel.
\newblock Acquisition and use of shortcut strategies by traditionally schooled children.
\newblock \emph{Educational Studies in Mathematics}, 71\penalty0 (1):\penalty0 1--17, 2009.
\newblock URL \url{https://doi.org/10.1007/s10649-008-9155-z}.

\bibitem[Tyen et~al.(2024)Tyen, Mansoor, Carbune, Chen, and Mak]{tyen-etal-2024-llms}
Gladys Tyen, Hassan Mansoor, Victor Carbune, Peter Chen, and Tony Mak.
\newblock {LLM}s cannot find reasoning errors, but can correct them given the error location.
\newblock In \emph{Findings of the Association for Computational Linguistics ACL 2024}. Association for Computational Linguistics, August 2024.
\newblock URL \url{https://aclanthology.org/2024.findings-acl.826}.

\bibitem[Unslothai(2023)]{unslothai_unsloth}
Unslothai.
\newblock Unsloth.
\newblock \url{https://github.com/unslothai/unsloth}, 2023.
\newblock URL \url{https://github.com/unslothai/unsloth}.
\newblock GitHub repository.

\bibitem[Wei et~al.(2022)Wei, Wang, Schuurmans, Bosma, brian ichter, Xia, Chi, Le, and Zhou]{cot}
Jason Wei, Xuezhi Wang, Dale Schuurmans, Maarten Bosma, brian ichter, Fei Xia, Ed~H. Chi, Quoc~V Le, and Denny Zhou.
\newblock Chain of thought prompting elicits reasoning in large language models.
\newblock In Alice~H. Oh, Alekh Agarwal, Danielle Belgrave, and Kyunghyun Cho, editors, \emph{Advances in Neural Information Processing Systems}, 2022.
\newblock URL \url{https://openreview.net/forum?id=_VjQlMeSB_J}.

\bibitem[Welleck et~al.(2023)Welleck, Lu, West, Brahman, Shen, Khashabi, and Choi]{welleck2022generating}
Sean Welleck, Ximing Lu, Peter West, Faeze Brahman, Tianxiao Shen, Daniel Khashabi, and Yejin Choi.
\newblock Generating sequences by learning to self-correct.
\newblock In \emph{The Eleventh International Conference on Learning Representations}, 2023.
\newblock URL \url{https://openreview.net/forum?id=hH36JeQZDaO}.

\bibitem[Yang et~al.(2024)Yang, Yang, Hui, Zheng, Yu, Zhou, Li, Li, Liu, Huang, et~al.]{yang2024qwen2}
An~Yang, Baosong Yang, Binyuan Hui, Bo~Zheng, Bowen Yu, Chang Zhou, Chengpeng Li, Chengyuan Li, Dayiheng Liu, Fei Huang, et~al.
\newblock Qwen2 technical report.
\newblock \emph{arXiv preprint}, 2024.
\newblock URL \url{https://arxiv.org/abs/2407.10671}.

\bibitem[Yarowsky(1995)]{yarowsky-1995-unsupervised}
David Yarowsky.
\newblock Unsupervised word sense disambiguation rivaling supervised methods.
\newblock In \emph{33rd Annual Meeting of the Association for Computational Linguistics}, pages 189--196, Cambridge, Massachusetts, USA, June 1995. Association for Computational Linguistics.
\newblock \doi{10.3115/981658.981684}.
\newblock URL \url{https://aclanthology.org/P95-1026}.

\bibitem[Ye et~al.(2023)Ye, Jo, Kim, Kim, Hwang, and Seo]{ye2023selfee}
Seonghyeon Ye, Yongrae Jo, Doyoung Kim, Sungdong Kim, Hyeonbin Hwang, and Minjoon Seo.
\newblock Selfee: Iterative self-revising llm empowered by self-feedback generation.
\newblock \emph{Blog post}, 2023.
\newblock URL \url{https://lklab.kaist.ac.kr/SelFee/}.

\bibitem[Yu et~al.(2024)Yu, Jiang, Shi, YU, Liu, Zhang, Kwok, Li, Weller, and Liu]{yu2024metamath}
Longhui Yu, Weisen Jiang, Han Shi, Jincheng YU, Zhengying Liu, Yu~Zhang, James Kwok, Zhenguo Li, Adrian Weller, and Weiyang Liu.
\newblock Metamath: Bootstrap your own mathematical questions for large language models.
\newblock In \emph{The Twelfth International Conference on Learning Representations}, 2024.
\newblock URL \url{https://openreview.net/forum?id=N8N0hgNDRt}.

\bibitem[Zhou et~al.(2023)Zhou, Sch{\"a}rli, Hou, Wei, Scales, Wang, Schuurmans, Cui, Bousquet, Le, and Chi]{l2m}
Denny Zhou, Nathanael Sch{\"a}rli, Le~Hou, Jason Wei, Nathan Scales, Xuezhi Wang, Dale Schuurmans, Claire Cui, Olivier Bousquet, Quoc~V Le, and Ed~H. Chi.
\newblock Least-to-most prompting enables complex reasoning in large language models.
\newblock In \emph{The Eleventh International Conference on Learning Representations}, 2023.
\newblock URL \url{https://openreview.net/forum?id=WZH7099tgfM}.

\end{thebibliography}

\newpage

\section{Appendix}

\subsection{Prompts}
\label{appendix:prompts}
\begin{figure*}[h!]
\small
% \sffamily
\centering
\begin{tabular}{l}
\toprule
\addlinespace
\parbox{\textwidth}{Below is an instruction that describes a task, paired with an input that provides further context. Write a response that appropriately completes the request.
\newline
\newline
Instruction:
Solve the given math problem step by step. Put your final answer after 'Final answer:'.
\newline
\newline
Input:
John buys 3 dress shirts.  They sell for \$20 each. He also has to pay 10\% tax on everything. How much did he pay in total?
\newline
\newline
Response:
The shirts cost 3*\$20=\$$<<$3*20=60$>>$60 before tax
The tax cost \$60*.1=\$$<<$60*.1=6\>\>6
So in total they paid \$60+\$6=\$$<<$60+6=66\>\>66
Final Answer: 66$<$eos$>$} \\
\addlinespace
\addlinespace
\parbox{\textwidth}{\centering \bf  [7 more examples randomly sampled from the training set]}\\
\addlinespace
\addlinespace
\parbox{\textwidth}{
Input:
Thomas is training at the gym to prepare for a competition. He trained for 5 hours every day for a month (30 days). If he continues to train for the next 12 days, how many hours will he spend on training in total?
\newline
\newline
Response:
\textcolor{teal}{Model generated response ..}} \\
\bottomrule
\end{tabular}
\caption{8-shot Chain of Thought demonstration.}
\label{fig:prompt-8shot-cot}
\end{figure*}

\begin{figure*}[h!]
\small
% \sffamily
\centering
\begin{tabular}{l}
\toprule
\addlinespace
\parbox{\textwidth}{Below is an instruction that describes a task, paired with an input that provides further context. Write a response that appropriately completes the request.
\newline
\newline
Instruction:
Solve the given math problem by decomposing it into smaller, manageable sub-questions. Put your final answer after 'Final answer: '.
\newline
\newline
Input:
John buys 3 dress shirts.  They sell for \$20 each. He also has to pay 10\% tax on everything. How much did he pay in total?
\newline
\newline
Response:
Sub-question 1: How much did John spend on shirts? \\
Answer to Sub-question 1: 3*\$20=\$$<<$3*20=60$>>$60 \\
Sub-question 2: How much did John spend on tax? \\
Answer to Sub-question 2: \$60*.1=\$<<60*.1=6>>6 \\
Sub-question 3: How much did John spend in total? \\
Answer to Sub-question 3: \$60+\$6=\$$<<$60+6=66$>>$66 \\
Final Answer: 66$<$eos$>$} \\
\addlinespace
\addlinespace
\parbox{\textwidth}{\centering \bf  [7 more examples randomly sampled from the training set]}\\
\addlinespace
\addlinespace
\parbox{\textwidth}{
Input:
Thomas is training at the gym to prepare for a competition. He trained for 5 hours every day for a month (30 days). If he continues to train for the next 12 days, how many hours will he spend on training in total?
\newline
\newline
Response:
\textcolor{teal}{Model generated response ..}} \\
\bottomrule
\end{tabular}
\caption{8-shot Least to Most demonstration.}
\label{fig:prompt-8shot-l2m}
\end{figure*}

\begin{figure*}[h!]
\small
% \sffamily
\centering
\begin{tabular}{l}
\toprule
\addlinespace
\parbox{\textwidth}{Below is an instruction that describes a task, paired with an input that provides further context. Write a response that appropriately completes the request.
\newline
\newline
Instruction:
Solve the given math problem by writing a python program. Store your result as a variable named 'answer'.
\newline
\newline
Input:
John buys 3 dress shirts.  They sell for \$20 each. He also has to pay 10\% tax on everything. How much did he pay in total?
\newline
\newline
Response:
total\_shirts = 3 \\
cost\_of\_one\_shirt = 20 \\
total\_cost\_shirts = total\_shirts * cost\_of\_one\_shirt \\
tax\_rate = 0.1  \\
tax\_amount = tax\_rate * total\_cost\_shirts \\
total\_cost = total\_cost\_shirts + tax\_amount \\
answer = total\_cost $<$eos$>$} \\
\addlinespace
\addlinespace
\parbox{\textwidth}{\centering \bf  [7 more examples randomly sampled from the training set]}\\
\addlinespace
\addlinespace
\parbox{\textwidth}{
Input:
Thomas is training at the gym to prepare for a competition. He trained for 5 hours every day for a month (30 days). If he continues to train for the next 12 days, how many hours will he spend on training in total?
\newline
\newline
Response:
\textcolor{teal}{Model generated response ..}} \\
\bottomrule
\end{tabular}
\caption{8-shot Program of Thought demonstration.}
\label{fig:prompt-8shot-pot}
\end{figure*}

\end{document}